\documentclass[runningheads]{llncs}
\usepackage{graphicx}
\usepackage{makecell}
\usepackage{amsmath}
\usepackage{amsfonts}
\usepackage{url}

\usepackage{hyperref}

\usepackage[marginal]{footmisc}

\begin{document}
\title{An Effective System for Multi-format Information Extraction}
\titlerunning{An Effective System for Multi-format IE}
%
\author{Yaduo Liu$^\dagger$  \and
Longhui Zhang$^\dagger$     \and
Shujuan Yin  \and Xiaofeng Zhao \and Feiliang Ren$^*$}
\authorrunning{Y.Liu \& L.Zhang et al.}
%
\institute{School of Computer Science and Engineering, \\Northeastern University, Shenyang, 110169, China\\
\email{renfeiliang@cse.neu.edu.cn}}

\maketitle              
\footnote{$^\dagger$ These authors contribute equally to this research and are listed randomly. \\$*$ Corresponding author.}
\begin{abstract}
The multi-format information extraction task in the 2021 Language and Intelligence 
Challenge is designed to comprehensively evaluate information extraction from different dimensions. It consists of an multiple slots relation extraction subtask and two   event extraction subtasks that extract events from both sentence-level and document-level. Here we describe our system for this multi-format information extraction competition task. 
Specifically, for the  relation extraction subtask, we convert it  to a  traditional 
triple extraction task and design a voting based method that makes full use of existing models. For the sentence-level event extraction subtask, we convert it to a NER task and use a pointer labeling based method for extraction. Furthermore, considering the annotated trigger information may be helpful for    event extraction,  we design an   auxiliary trigger recognition model and  use the multi-task learning mechanism to  integrate the trigger features  into the event extraction model.  For the document-level event extraction subtask, we design an Encoder-Decoder based method and propose a   Transformer-alike  decoder. Finally, our system  ranks No.4 on the  test set leader-board of this multi-format information extraction task, and its F1 scores for the subtasks of relation extraction, event extractions of sentence-level and document-level   are 79.887\%,  85.179\%, and 70.828\% respectively. The codes of our model are available at {https://github.com/neukg/MultiIE}. 

\end{abstract}
\section{Introduction}

Information extraction (IE) aims to extract structured knowledge from 
unstructured texts. Named entity recognition (NER), relation extraction (RE) and event extraction (EE) are some  fundamental information extraction tasks that focus on extracting entities, relations and events respectively. However, most researches only focus on extracting information in a single format, while lacking a unified evaluation platform for IE in different formats. Thus the 2021 Language and Intelligence Challenge (LIC-2021) sets up a multi-format IE competition task which is  designed to comprehensively evaluate IE from different dimensions. The task consists of a relation extraction subtask and two event extraction subtasks that extract events from both sentence-level and document-level. The definitions of these subtasks are as follows$^1$\footnote{$^1$https://aistudio.baidu.com/aistudio/competition/detail/65}. 

\textbf{Relation Extraction} is a task that aims to extract all SPO triples from a given sentence according to a predefined schema set. The schema set defines relation P and the types of its corresponding subject S and object O.  According to the complexity of object O, there are two types of schemas. The first one is the single-O-value schema in which the object O gets only a single slot and value. The second one is the  schema in which the object O is a structure composed of multiple slots and their corresponding values.

\textbf{Event Extraction} consists of two subtasks. The first one is the sentence-level event (SEE) extraction whose aim is that given a sentence and predefined event types with corresponding argument roles, to  identify all events of target types mentioned in the sentence, and extract corresponding event arguments playing target roles. The predefined event types and argument roles restrict the scope of extraction. The second one is the document-level event (DEE) extraction task, which shares almost the same definition with the previous one, but considers input text fragments in document-level rather than sentence-level, and restricts the event types to be extracted in the financial field.

Table~\ref{tab1} demonstrates some schema examples for these subtasks. For example, in the given multiple-{O}-values schema RE  example, the object consists of two items: \emph{inWork} and \emph{@value}, but traditional RE may only get \emph{@value}
item in this  example.
Compared with traditional single-format IE task, there are following challenges in the  multi-format IE task designed in  LIC-2021. 

First,  despite the success of existing joint entity and relation extraction methods, they cannot be directly applied to the multiple slots RE task in LIC-2021 because   of the multiple-O-values schema relations.

\begin{table}[t]
	\caption{Schema examples of three information extraction subtasks in LIC-2021.}\label{tab1}
	\centering
	\begin{tabular}{|l|l|}
		\hline
		Task & Schema example \\
		\hline
		RE (multiple-O-values schema) & \makecell[l]{object\_type: \{inWork: film and television work, \\ @value: role\} \\ predicate: play \\ subject\_type: entertainer} \\
		\hline
		SEE & \makecell[l]{event\_type: \mbox{finance/trading-limit down} \\ role\_list: [role: \mbox{time}, role: \mbox{stocks fell to limit}]}\\
		\hline
		DEE & \makecell[l]{event\_type: \mbox{be interviewed} \\ role\_list: [role: \mbox{the name of company}, \\ role: \mbox{disclosure  time},\\role: \mbox{time of being interviewed},\\role: \mbox{interviewing institutions}]}\\
		\hline
	\end{tabular}
\end{table}

Second, the event  extraction is more challenging due to the \emph{overlapping event role} issue: an argument may have multiple roles. 
Besides, there are annotated triggers provided in the given datasets, and how to effectively use this kind of  trigger information is still an  open issue. 


Third, it is difficult  to extract different events that have the  same event type in the DEE subtask. 
For example, for the ``\emph{be interviewed}'' event type shown in Table~\ref{tab1},  different people maybe interviewed at different time. For this case, in the SEE subtask, two different time arguments and two interviewed people are regarded as arguments for the same event. However, these two arguments need to be classified into arguments of two different events in the DEE subtask.

In our system, we use some effective methods to overcome these challenges. Specifically, for the first one, we design a schema disintegration module to convert   each multiple-O-values relation into several single-O-value relations, then use a voting based module to obtain the final relations. 
For the second one, we convert the SEE subtask into a NER task and use a pointer labelling based method to address the mentioned \emph{overlapping} issue. To  use  the trigger information, we  design an auxiliary trigger recognition module and use the multi-task learning mechanism to integrate the trigger information. For the  third one, we design an Encoder-Decoder based method to \emph{generate} different events. 


\section{Related Work}
\label{sec:relwork}
RE is a long-standing natural language processing task whose aim is to mine factual triple knowledge from free texts. These triples have the form of (S, P, O), where both S and O are entities and P is a kind of semantic relation that connects S and O. For example, a triple (Beijing, capital\_of, China) can express the knowledge that \emph{Beijing is the capital of China}. 
Currently, the methods that extract entities and relations jointly in an end to end way are dominant.  According to the  extraction route taken, these joint extraction methods can be roughly classified into following three groups. (i) Table filling based
methods (\cite{gupta-etal-2016-table,wang-lu-2020-two}),
that maintain a table for each relation, and each item in such a table is used to  indicate whether the row word and the column word possess the corresponding relation.  (ii) Tagging based methods (\cite{hang-stel,wei-etal-2020-novel}) that  use the sequence labeling based methods to tag entities and relations.  (iii) Seq2seq based methods (\cite{zeng-etal-2018-extracting,DBLP:conf/aaai/ZengZL20}),
that use the sequence generation based methods to \emph{generate} the items in each triple with a predefined order, for example,  first \emph{generates} S, then P, then O, etc.

Event extraction (EE)  can be categorized into sentence-level EE (SEE) and document-level EE (DEE)  according to the input text. 
The majority of existing EE researches (such as~\cite{chen-etal-2015-event} and  ~\cite{wang-etal-2019-adversarial-training}) focus on SEE.
Usually, they use a pipeline based framework that predicts triggers firstly,  and then predict arguments. 
Recently, DEE is  attracting more and more attentions. Usually there are two main  challenges in DEE, which are: (i)  arguments of one event may scatter across some long-distance sentences of a document or even different documents, and (ii) each document is likely to contain multiple events. A representative Chinese DEE work is \cite{yang-dcfee-2018}, whose DEE model contains two main components: a SEE model that extracts event arguments and event triggers from a sentence;  and a DEE model that extracts event arguments from the whole document based on a key event detection model and an arguments-completion strategy. 
Another representative Chinese DEE work is 
\cite{zheng-etal-2019-doc2edag}, which uses
an entity-based directed acyclic graph to fulfill the DEE task  effectively. Moreover, they redefine a DEE task with the no-trigger-words design to ease document-level event labeling. Both of these two DEE work are evaluated on Chinese financial field. 

\begin{figure}[t]
\centering
\includegraphics[width=\textwidth, keepaspectratio]{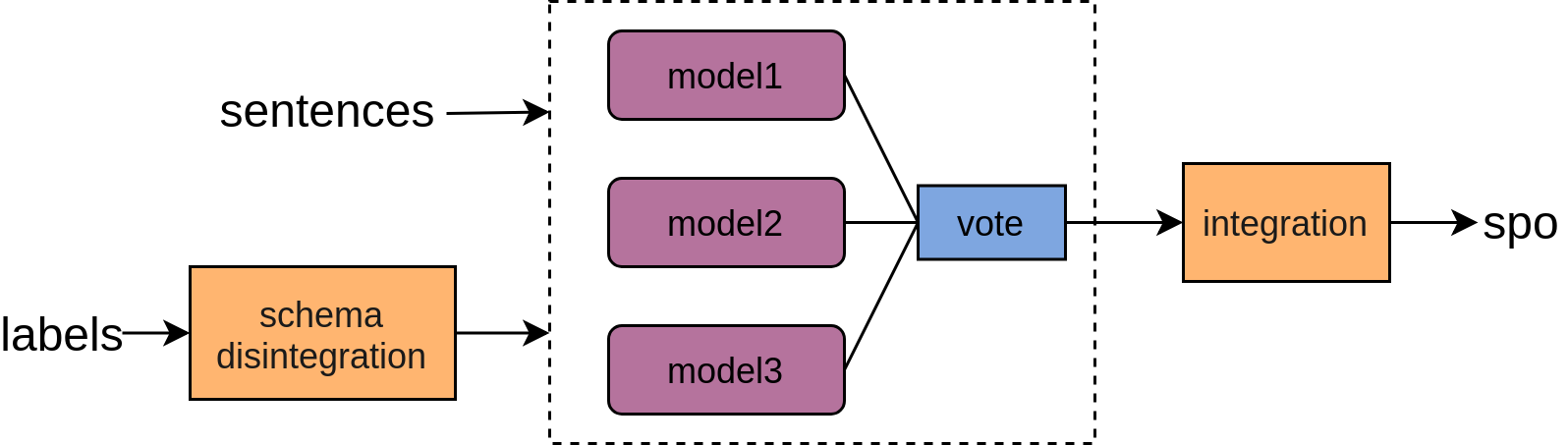}
\caption{The architecture of our RE model.} \label{fig1}
\end{figure}
\section{Methodology}
\label{sec:method}
\subsection{Relation Extraction}

The architecture of our RE  model is shown in Fig~\ref{fig1}. We can see that it consists of  two main modules:   a schema disintegration module and a voting module. 

Given a multiple-O-values triple$^2$\footnote{$^2$Although a multiple-O-values relation contains multiple objects,  for simplicity, we still call it as a  triple.} \{s, p, [($o_{k_1}$, $o_{v_1}$), ..., ($o_{k_m}$, $o_{v_m}$)]\},
the schema disintegration module  would   transform it into \emph{2m-1} single-O-value triples. The concrete transformation process is as follows. First, $o_{v_1}$ will be taken as an object to form a triple \{s, p, $o_{v_1}$\}. Then from $i = 2$  on, each ($o_{k_i}$, $o_{v_i}$) will form following two triples: \{s, p-$o_{k_i}$, $o_{v_i}$\} and \{$o_{v_1}$, $o_{k_i}$, $o_{v_i}$\}. Here both $s$ and $o_{v_1}$ will be repeatedly taken as subjects in the formed triples, and p-$o_{k_i}$ is a new generated predicate. Accordingly, the given example will generate following triples,  which are \{s, p, $o_{v_1}$\}, \{s, p-$o_{k_2}$, $o_{v_2}$\}, \{$o_{v_1}$, $o_{k_2}$, $o_{v_2}$\}, ..., \{s, p-$o_{k_m}$, $o_{v_m}$\}, \{$o_{v_1}$, $o_{k_m}$, $o_{v_m}$\}. If there are some single-O-value triples, they can be formed into a  multi-O-value triple with a reverse process. 

In the voting module, three existing state-of-the-art triple extraction models are used to  extract single-O-value triples separately. Then their results would be voted to output the  triples that receive more votes. Next,  these obtained triples are  converted into some multiple-O-values triples$^3$\footnote{$^3$Note the single-O-value triples can also be viewed as a specifical kind of multi-O-value triples.} as final results of this RE task.  In our system,   following three existing state-of-the-art models are used for voting:  \emph{TPLinker}~\cite{wang-etal-2020-tplinker}, \emph{SPN}~\cite{2020arXiv201101675S} and \emph{CasRel}~\cite{wei-etal-2020-novel}. 

\subsection{Sentence-level Event Extraction}
\label{SEE}

In our system, we treat each event argument  as an entity, and concatenate this argument's corresponding event type and role as the entity type. Then the SEE task is converted into a NER task. 
But there is a \emph{multiple label} issue in the SEE task: an argument may belong to multiple event types. To address this issue, here we use a pointer labeling based NER method that   tags the start token and end token of an entity span for each candidate entity type.  Specifically, as shown in Eq.~\eqref{eq:1}, for each entity type, the model compute two probabilities for each word to indicate the possibilities of this word being the start and end tokens of an entity that possesses this entity type. 
\begin{equation}
\begin{aligned}
&p_{s_{ij}} = sigmoid(W_i^Sh_j + b_i^S), \\
&p_{e_{ij}} = sigmoid(W_i^Eh_j + b_i^E)
\label{eq:1}
\end{aligned}
\end{equation}
where $\{W_i^{(.)}\in\mathbb{R}^{d}\}_{i=1}^r$, $\{b_i^{(.)}\in\mathbb{R}\}_{i=1}^r$ are learnable parameters for the $i$-th entity type, $h_j\in\mathbb{R}^{d}$ is the token representation for the $j$-th word and is  obtained by a pretrained language model,
$r$ is the number of entity types, $p_{s_{ij}}$ and $p_{e_{ij}}$ is the probabilities of the $j$-th word being the start token and the end token of an entiy that should be labeled with the $i$-th entity type.

Furthermore, to make full use of the features from the annotated triggers, we design an auxiliary trigger recognition module that also recognizes triggers in the same NER manner as used above.  This auxiliary module  are trained jointly with the above argument recognition module in a multi-task learning manner. 

During training, the trigger recognition module will generate  a representation for each trigger. Then these  trigger representations will be  merged into the token representations used by the argument recognition module with an attention mechanism for the argument recognition of next iteration. 
But during inferencing,  our model will first recognize all triggers (the results are denoted as $T = \{t_1, t_2, ...\}$). Then these triggers' representations will be fused into a unified representation (denoted as $t$) by a max-pooling operation. Next,  an additive-attention based method is used to obtained a new token representation sequence (denoted as $\hat{H}$), which will then be used as input for the argument recognition module. Specifically,  $\hat{H}$ is computed with following  Eq.~\eqref{eq:2}.
\begin{equation}
\begin{aligned}
&\alpha = softmax(V^\mathrm{T}tanh(W_1H + W_2t)), \\
&\hat{H} = \alpha \cdot H 
\label{eq:2}
\end{aligned}  
\end{equation}
where $H$ is the original token representations, $V$, $W_1$, and $W_2$
are learnable parameters, and the superscript $\mathrm{T}$ denotes a matrix transpose operation. 

\begin{figure}[t]
	\centering
	\includegraphics[width=0.9\textwidth, keepaspectratio]{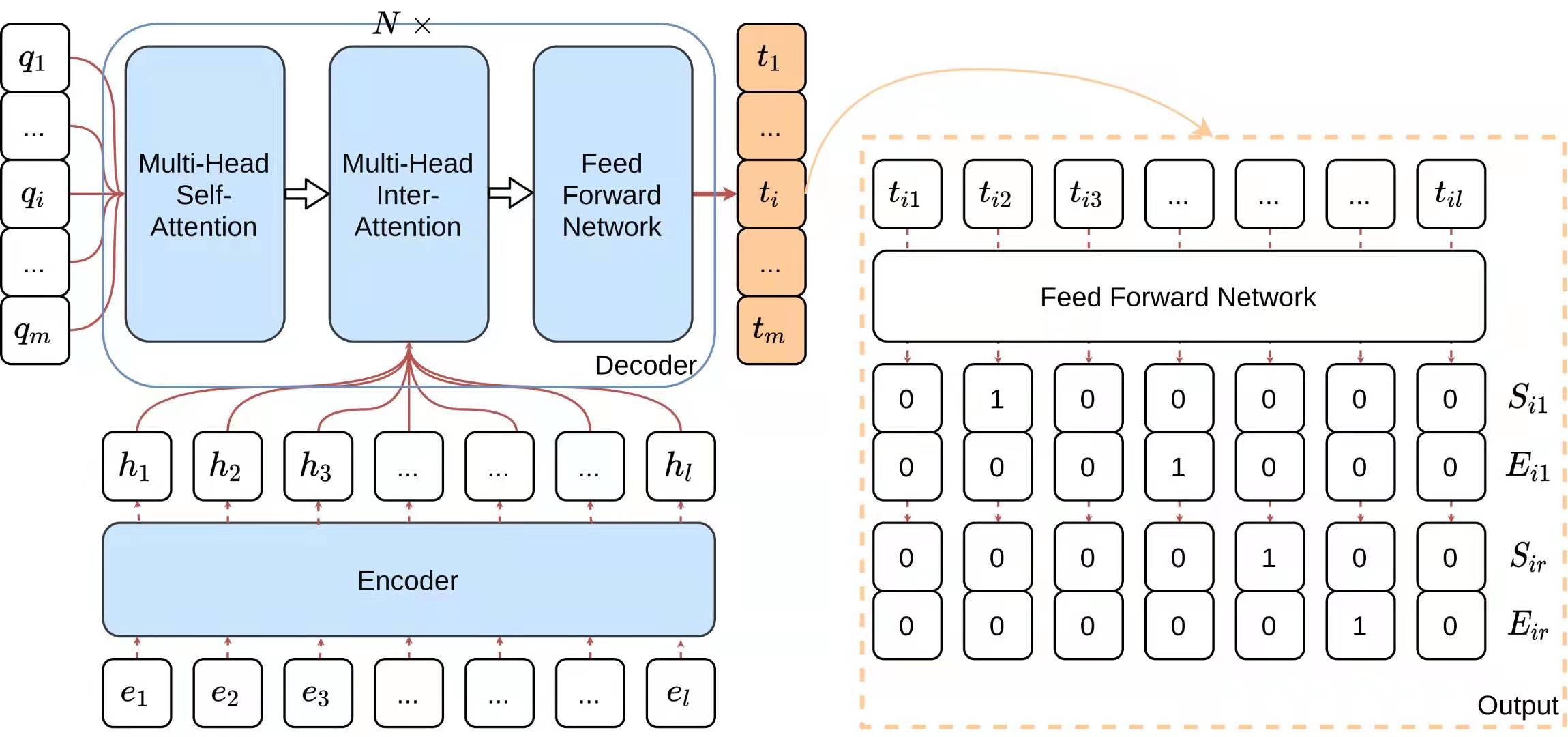}
	\caption{The  architecture of our  DEE model.  \emph{N} is the number of transformer decoder. 
	} \label{fig2}
\end{figure}

\subsection{Document-level Event Extraction}

The architecture of our DEE model is shown  in Fig.~\ref{fig2}. We can  see it contains three  main modules, which are \emph{Encoder},  {Decoder}, and  \emph{Output}. 


BERT~\cite{devlin-etal-2019-bert}  and some of its variants are used as    \emph{Encoder}  to get a context-aware representation for each token in a document. Note that both BERT and its variants have a length restriction for the input text (usually 512 TOKENs),  so we use the \emph {sliding window} based method to split a long document into several segments. Then  each segment is taken as an input to be processed by our DEE model and the results of these segments  are combined as final results.  Given a segment $S$ = $\{e_0, e_1, ..., e_l\}$, the output of \emph{Encoder} is a token   embedding sequence, we  denote it  as $H \in \mathbb{R}^{l \times d}$, where $l$
is the number of tokens in the given segment,  and $d$ is the dimension of the token embedding.


In the \emph{Decoder} module, some learnable query embeddings (as shown  in Fig.\ref{fig2}, denoted as $({q_0, q_1,...,q_m})$) along with the token embedding sequence outputted by the \emph{Encoder} module are taken as input to predict events. Each query embedding corresponds to a specific event. Although different documents may have different numbers of events, we set a fixed number (denoted as $m$, and is set to 16 here) for the used queries, which  is obtained according to the maximal  event number in a document of  the training dataset.  

Here our designed  \emph{Decoder}  is similar to the one used in Transformer~\cite{NIPS2017_3f5ee243}, but we delete the components of  masking and  positional encoding because  there is no correlations and position relations between event's query embeddings in the DEE task. Our \emph{Decoder} module consists of \emph{N} stacked layers, and each layer consists of a multi-head self-attention module, a multi-head inter-attention module, and a feed forward network. It  will generate a refined embedding representation sequence (denoted as $T \in \mathbb{R}^{m\times l\times d}$) for the input queries.  Specifically,  the multi-head self-attention module  is first  performed to highlight some queries. 
Then, the multi-head inter-attention operation is performed  to highlight the correlations between  this input queries  and the input sentence.
Next, a feed forward network is used to generate a new embedding representation sequence for the input query. 
The above three steps will be  performed \emph{N} times to obtained  $T$. 

The \emph{Output} module  is just the same as the  extraction model   for   DEE, which uses a pointer labeling based method and  an auxiliary trigger recognition module together to predict the event arguments. Specifically, it takes  each $t_i \in \mathbb{R}^{ l\times d}$ in $T$ as  input, and  computes two kinds of probabilities, $S_i \in \mathbb{R}^{l \times r}$ and $E_i \in \mathbb{R}^{l \times r}$, to denote the probabilities of each word  being the 
start  and end tokens of an entity that should be labeled with the $i$-th event argument.



\subsection{Loss function} 
\label{section:loss}
The cross entropy loss function is used for both the tasks of RE and SEE,  but we use a bipartite matching loss for DEE. The main difficulty in DEE is that 
the predicted arguments are not always in the same order as those in the ground truths. For example, an event is predicted by the $i$-th query embedding,
but it may be in the $j$-th position  of the ground truths. Thus, we do not apply the  cross-entropy loss function  directly because this  loss is sensitive
to the permutation of the predictions. Instead, we use the bipartite matching loss proposed in \emph{SPN} to produce an optimal matching between $m$ predicted events and $m$ ground truth events.
Specifically, we concatenate  $S$ and $E$ (see above,   generated in DEE),  into $\hat{Y} \in \mathbb{R}^{m \times l \times 2 \times r}$,  that can represent the predicted events. The ground truths is denoted as $Y \in \mathbb{R}^{m \times l \times 2 \times r}$. The process of computing bipartite matching loss
is divided into two steps: finding an optimal matching and computing the loss function. To find optimal matching between $Y$ and $\hat{Y}$, 
we search for a permutation of elements $\pi^*$ with the lowest cost $l_{match}$, as shown in Eq.\eqref{eq:3}. 
\begin{equation}
\pi^* = \mathop{\arg\min}_{\pi \in S(m)} \sum_{i=1}^m l_{match}(Y_i, \hat{Y}_{\pi{(i)}})
\label{eq:3}
\end{equation}
where $S(m)$ is the space of all $m$-length permutations. $Y_i \in \mathbb{R}^{l \times 2 \times r}$ and $\hat{Y}_{\pi{(i)}} \in \mathbb{R}^{l \times 2 \times r}$ are outputs of the $i$-th predicated event and the $i$-th ground truth.
$l_{match}(Y_i, \hat{Y}_{\pi{(i)}})$ is the pair-wise  matching cost between the ground truth $Y_i$ and the prediction event with index ${\pi{(i)}}$, and it is be computed as follow:
\begin{equation}
l_{match} = \hat{Y}_{\pi(i)} \circ  Y_i
\end{equation}
where $\circ $ is Hadamard product, and the optimal $\pi^*$ can be be computed in polynomial time (\emph{$O(m_3)$}) via the Hungarian algorithm$^4$\footnote{$^4$https://en.wikipedia.org/wiki/Hungarian algorithm}. Readers can find more detailed introduction about this loss in \emph{SPN}. 

After $\pi^*$ is obtained, we change the event order of the predicated $\hat{Y}_{\pi^*}$ to be in line with $\pi^*$,  and the re-ordered result is denoted as  $Y^*$.
Then we use the cross entropy loss function  to compute the loss between $Y^*$ and $Y$.

\subsection{Model Enhancement Techniques}

In our system, some model enhancement techiques are used to further improve the performance of each subtask. 

\noindent\textbf{Adversarial Training} To train a robust model, we  use the  FGM adversarial training mechanism~\cite{DBLP:conf/iclr/MiyatoDG17} to disturb the direction of  gradients, which makes the loss  increase toward the maximum direction. Then  the model will be pushed to find more robust parameters during the training process to weaken the impact of this aleatoric uncertainty. 
Assume the input text's embedding representation sequence is $x$, then its disturbance  embedding $r_{adv}$ is computed with Eq.\eqref{eq:5}. 
\begin{equation}
\begin{aligned}
&g = \nabla_x L(\theta, x, y), \\ 
&r_{adv} = \epsilon \cdot g/\|g\|_2
\label{eq:5}
\end{aligned}
\end{equation}
where $\epsilon$ is a hyperparameter to control the disturbance degree. 

During training, $r_{adv}$ will be  updated with  above equation after the loss of a subtask is computed. Then it  will be taken as  a new  input of each task for the training of  next  iteration. 

\noindent\textbf{Data Augmentation} We  use following two kinds of  data augmentation strategies to enhance the performance of our models for different tasks. The first strategy is \emph{Synonyms Replace}, which first randomly selects some words from each input text,  and then these words will be replaced by their synonyms that are  selected from a synonym dictionary. The second strategy is \emph{Randomly Delete}, under which  every input word is randomly deleted with a predefined probability. But if a word to be deleted is an entity, it will be remained.

\noindent\textbf{Model Ensemble} 
We use the bagging strategy   to ensemble multiple base models. It simply average or vote the  weighted  results  of some  base models. Concretely, we split all data in the training set into 10 folds, and train with 9 folds and validate with the remained 1 fold. By this way, we can obtain  ten different base models.  Besides,  we also replace the \emph{Encoder} of different models with different pretrained language models, including BERT, RoBERTa-wwm-ext~\cite{2019arXiv190711692L}, and NEZHA~\cite{2019arXiv190900204W}. Accordingly,  another kinds of base models can be trained. Finally,  all the  base models of a task  are ensembled into one model.  In our system, we simply vote on the predicted results of all base models, and select the outputs that get more votes.

\begin{table}[t]
	\caption{Statistics of dataset.}\label{tab:stas}
	\centering
	\begin{tabular}{|l|l|l|l|}
		\hline
		Dataset & Training Set & Development Set & Test Set \\
		\hline
		DuIE2.0 & 171,293 & 20,674 &21,080\\
		\hline
		DuEE1.0 & 11,958 & 1,498 &3,500\\
		\hline
		DuEE-fin & 7,047 & 1,174 &3,524\\
		\hline
	\end{tabular}
\end{table}

\section{Experiments}
\label{sec:experiments}
\subsection{Basic Settings}
\textbf{Datasets} The LIC-2021 competition uses three large-scale Chinese information extraction datasets, including DuIE2.0~\cite{li2019duie}, DuEE1.0~\cite{li2020duee} and DuEE-fin~\cite{li2020duee}.  

DuIE2.0 is the largest schema-based Chinese relation extraction dataset, which contains more than 430,000 triples over 210,000 real-world Chinese sentences, restricted by a pre-defined schema of 48 relation types. In this dataset, there are 43 single-\emph{O}-value relations and 5 multiple-\emph{O}-values relations.
DuEE1.0 is a Chinese event extraction dataset, which consists of 17,000 sentences containing 20,000 events of 65 event types.
DuEE-fin is a document-level event extraction dataset in the financial domain. The dataset consists of 11,700 documents across 13 event types, in which there are negative samples that do not contain any target events. All of these datasets are  built by Baidu. Some basic statistics of these datasets are shown in Table \ref{tab:stas}. 

\noindent\textbf{Evaluation Metrics}
\label{sec:evaluation}
According to the settings of LIC-2021,  the scores of a participating system on DuIE2.0, DuEE1.0 and DuEE-fin are given respectively, and the macro average is used as the final score of the system.  F1 score is used as the evaluation metrics for all three subtasks.
For the RE subtask,  a predicted relation with multiple-O-values schema would be regarded as correct only when all its slots are exactly matched with a manually annotated golden relation. 
For the SEE subtask, a predicted event argument  is evaluated according to  a token level matching F1 score. If an event argument has multiple annotated mentions, the one with the highest matching F1 will be used. 
For the DEE subtask, for each document,  the evaluation would will first select a most similar predicted event for every annotated event (Event-Level-Matching). The matched predicted event must have the same event type as the annotated event and predict the most correct arguments. Each predicted event is matched only once.  

\begin{table}[t]
\caption{Models' performance in relation extraction, sentence-level event extraction and document-level event extraction.}\label{tab3}
\centering
\begin{tabular}{|l|l|l|l|l|l|}
\hline
RE & F1 score & SEE & F1 score & DEE & F1 score \\
\hline
CasRel & 0.7568 & Backbone & 0.8254 & Backbone & 0.6630\\
TPLinker & 0.7605 & Backbone+TR & 0.8402 & Backbone+TR & 0.6843\\
SPN & 0.7324 & & & Backbone+BML & 0.6702\\
\hline
Ens & 0.7894 & Ens & 0.8509 & Ens & 0.7089\\ 
Ens+AT & 0.7905 & Ens+AT & 0.8535 & Ens+AT & 0.7134\\
Ens+AT+PLM & 0.8051 & Ens+AT+PLM & 0.8594 & Ens+AT+PLM & 0.7235\\
\hline
\end{tabular}
\end{table}

\noindent\textbf{Implementation Details}
AdamW~\cite{2017arXiv171105101L} is used to train all the models . The learning rates for RE, SEE, and DEE are 3e-5, 5e-5, and 5e-5 respectively. And the epochs for these three subtasks are 10, 40, and 30 respectively. 

\subsection{Results}
\textbf{Main Results} The main experimental results are shown in Table~\ref{tab3}, in which Backbone, TR, BML,  Ens, AT, and PLM  to denote the backbone models, the trigger recognition module,  the bipartite matching loss,  the ensemble model,  the adversarial training, 
and the pretrained language models respectively.
In all of our experiments, Backbone refers to the model in which the transformer decoder and the  bipartite matching loss are removed. 

From Table~\ref{tab3} we can see that our RE model achieves far better results than all the compared state-of-the-art triple extraction models like \emph{CasRel}, \emph{TPLinker}, and \emph{SPN}. Here we donot compare our two EE models  with other state-of-the-art models because  most of existing EE models cannot be used here directly. 


\noindent\textbf{Ablation Results}
From the ablation results in Table \ref{tab3} we can see that  the pretrained language models  are much helpful for all subtasks and they always bring a significant performance gain for a subtask. Besides, the adversarial training   is also helpful and it consistently improves the performance of a subtask.


From the results of both SEE and DEE we can see that the designed  trigger recognition module plays a very important role to the performance, and it improves nearly 1.5\% point for the F1 score of   SEE, and more than 2.1\% point for the F1 score of  DEE. In fact, this module plays even more roles than the pretrained language models.  Based on these results we can conclude that  the triggers do contain some important cues for both kinds of EE subtasks, and making full of these cues are much helpful for improving  the performance. 

From the results of DEE we can see that the bipartite matching loss also plays a helpful role to improve the performance. And it brings more performance gain than the adversarial training.

Besides, we also conduct  experiments on the DEE subtask to evaluate: (i) the impact of the layer number of the transformer-based  decoder (the number $N$ in Fig.\ref{fig2}) , and (ii) the impact of different sizes of sliding windows. For the first one,  we test different numbers in the range of 0 to 5. Our experiments show that  when the number of decoder layer is set to 3, the model achieves the best results. 
We think this is mainly because that a moderate layer number will  lead to more complete integration of input information into event queries, while a larger or smaller number will make the model be overfit or underfit. 
For the second one, we set the size of sliding windows  to 128, 256, and 512. And their  F1 scores   are 64.8\%, 66.9\%, and 69.2\% respectively when the number of the decoder's layer is set to 3 and the bipartite matching loss is used. These results show that usually, the performance would increase as the size of the sliding window increases. This is because that more  context features can be taken into consideration with a larger  sliding windows size. 

\section{Conclusions}
\label{sec:conclusions}
This paper describes our system to the LIC-2021 multi-format IE task. We use different methods to overcome different challenges in this competition task, including the  schema disintegration method for the multiple-O-values schema issue in the RE subtask,  the multi-task learning method for  the SEE subtask, and the Transform-alike decoder  for the DEE subtask. Experimental results show that our system is effective and it ranks No.4 on the final test set leader-board of this competition. Its F1 scores are  79.887\% on  DuIE2.0, 85.179\% on DuEE1.0, and 70.828\% on DuEE-fin respectively.

However,  there is still plenty of room for improvement, and lots of work should be further explored.  First, in the RE subtask,
many triples are not annotated. These triples gives model wrong supervisions, which is very harmful for the performance. But this \emph{missing annotation} issue is still an open issue, and should be further explored.  Second, in the DEE subtask, how to process long text is still a challenge that is worthy being further studied. In addition, if two arguments of one event are far away in the given text (either a sentence or a document), it would be difficult to extract them correctly. This issue also should be well  studied in the future. 

\section*{Acknowledgments}
	This work is supported by the National Key R\&D Program of China (No.2018YF
	C0830701), the National Natural Science Foundation of China (No.61572120), the Fundamental Research Funds for the Central Universities (No.N181602013 and No.N171602003), Ten Thousand Talent Program (No.ZX20200035), and Liaoning Distinguished Professor (No.XLYC1902057).
%
%
%
%
%
%
\bibliographystyle{splncs04}
\bibliography{references}

\end{document}